\documentclass{article}
\usepackage{ijcai20}
\usepackage{times}
\usepackage{soul}
\usepackage{url}
\usepackage{array}
\usepackage{booktabs}
\usepackage{multirow,multicol}
\usepackage[hidelinks]{hyperref}
\usepackage[utf8]{inputenc}
\usepackage[small]{caption}
\usepackage{graphicx}
\usepackage{amsmath}
\usepackage{amssymb}  
\usepackage{amsthm}
\usepackage{booktabs}
\usepackage{algorithm}
\usepackage{algorithmic}
\newcommand{\R}{{\mathbb R}}

\newcommand{\be}{\begin{eqnarray}}
\newcommand{\ben}{\begin{eqnarray*}}
\newcommand{\en}{\end{eqnarray}}
\newcommand{\enn}{\end{eqnarray*}}

\title{Robust Trajectory Forecasting for Multiple Intelligent Agents 
in Dynamic Scene}

\author{
Yanliang Zhu$^1$
\and
Dongchun Ren$^1$\and
Mingyu Fan$^{1,2}$\and
Deheng Qian$^1$ \and
Xin Li$^1$ \and
Huaxia Xia$^1$
\affiliations
$^1$Meituan-Dianping Group, Beijing 100102 China\\
$^2$School of Computer Science, Wenzhou university, Wenzhou 325035, China
\emails
\{zhuyanliang, rendongchun\}@meituan.com,
fanmingyu@wzu.edu.cn,
\{qiandeheng, lixin125, xiahuaxia\}@meituan.com
}

\begin{document}
\maketitle

\begin{abstract}
Trajectory forecasting, or trajectory prediction, of multiple interacting agents in dynamic scenes, is an important problem for many applications, such as robotic systems and autonomous driving. The problem is a great challenge because of the complex interactions among the agents and their interactions with the surrounding scenes. In this paper, we present a novel method for the robust trajectory forecasting of multiple intelligent agents in dynamic scenes. The proposed method consists of three major interrelated components: an interaction net for global spatiotemporal interactive feature extraction, an environment net for decoding dynamic scenes (i.e., the surrounding road topology of an agent), and a prediction net that combines the spatiotemporal feature, the scene feature, the past trajectories of agents and some random noise for the robust trajectory prediction of agents. Experiments on pedestrian-walking and vehicle-pedestrian heterogeneous datasets demonstrate that the proposed method outperforms the state-of-the-art prediction methods in terms of prediction accuracy.
\end{abstract}

\section{Introduction}

Trajectory  prediction of agents in dynamic scene is a challenging and essential task in many  fields, such as social-aware robotic systems \cite{nbody2011}, autonomous driving \cite{baiduAAAI2019} and behavior understanding \cite{peekingfuture2019}. Intelligent agents, such as humans, vehicles, and independent robots, are supposed to be able to understand and forecast the movement of the others to avoid collisions and make smarter movement plans.  Trajectory prediction has been studied extensively.  Traditional prediction methods, such as the Gaussian process regression \cite{GP2005}, the kinematic and dynamic method \cite{IMM2009} and the Bayesian networks method \cite{bayesian2011}, ignore the interactions among the agents and are only able to make short term predictions.  Recently, Recurrent Neural Network (RNN) and its variants \cite{socialLSTM2016}, such as Long Short-Term Memory (LSTM) and Gated Recurrent Unit (GRU), have shown promising ability in capturing the dynamic interactions of agents and a great number of trajectory prediction methods have been proposed based on them. 

However, trajectory prediction is still a challenging task because of the several properties of it:  1) When intelligent agents move in public, they often interact with other agents such as human or obstacles in the scene, which is named as the {\bf social behavior}. Actions, including collision avoidance and moving in groups, require the ability to forecast the possible movements or actions of the other agents. The social interactions may not be confined to nearby agents or obstacles.  2) The movement of agents is not only dependent on the nearby agents, but is also influenced by the surrounding physical scene, i.e., the {\bf dynamic scene}. One important factor of the scene is the road topology, such as intersections, turns, and slip lanes. Certain road topology can significantly influence the speed and direction of the moving agents.  An autonomous agent should be always moving on a feasible terrain.  3) The {\bf multi-modal motion} property illustrates that the interactive agents may follow several viable trajectories as there is a rich choice of reasonable movements. When two independent agents move toward each other, there are many possible different future trajectories that could avoid collision, such as moving to the left, to the right, or stop.

In this study, we propose a novel robust trajectory forecasting method for multiple intelligent agents in dynamic scene. The main contributions of this paper are summarized as follows.
\begin{itemize}
\item We model the global spatio-temporal interaction through an interaction net with a soft agent-tracking module. The interaction net not only considers the current locations and interaction of the agents, but also the temporal interactions among the agents by the hidden states of LSTMs on past trajectories.
\item  An environmental net is introduced to encode the dynamic scene. The surrounding road topology, such as interactions, turns and slip lanes, is firstly transformed into an high-definition map and then the map is encoded by a pre-trained convolutional neural network. 
\item Our trajectory prediction net combines the feature of spatio-temporal interaction, the environment feature, and the past trajectory to forecast the future trajectory of all the agent. Attention model is used to adaptively encode the spatio-temporal interaction of an agent with the others.
\end{itemize}

The rest of this paper is structured as follows: in Section 2, some related work is reviewed. The proposed robust trajectory prediction method is introduced in Section 3. Experimental comparisons with the state-of-the-art trajectory prediction methods on benchmark datasets are presented in Section 4. Finally, the conclusion is drawn in Section 5.

\begin{figure*}[thpb]
	\centering
	\includegraphics[width=14cm]{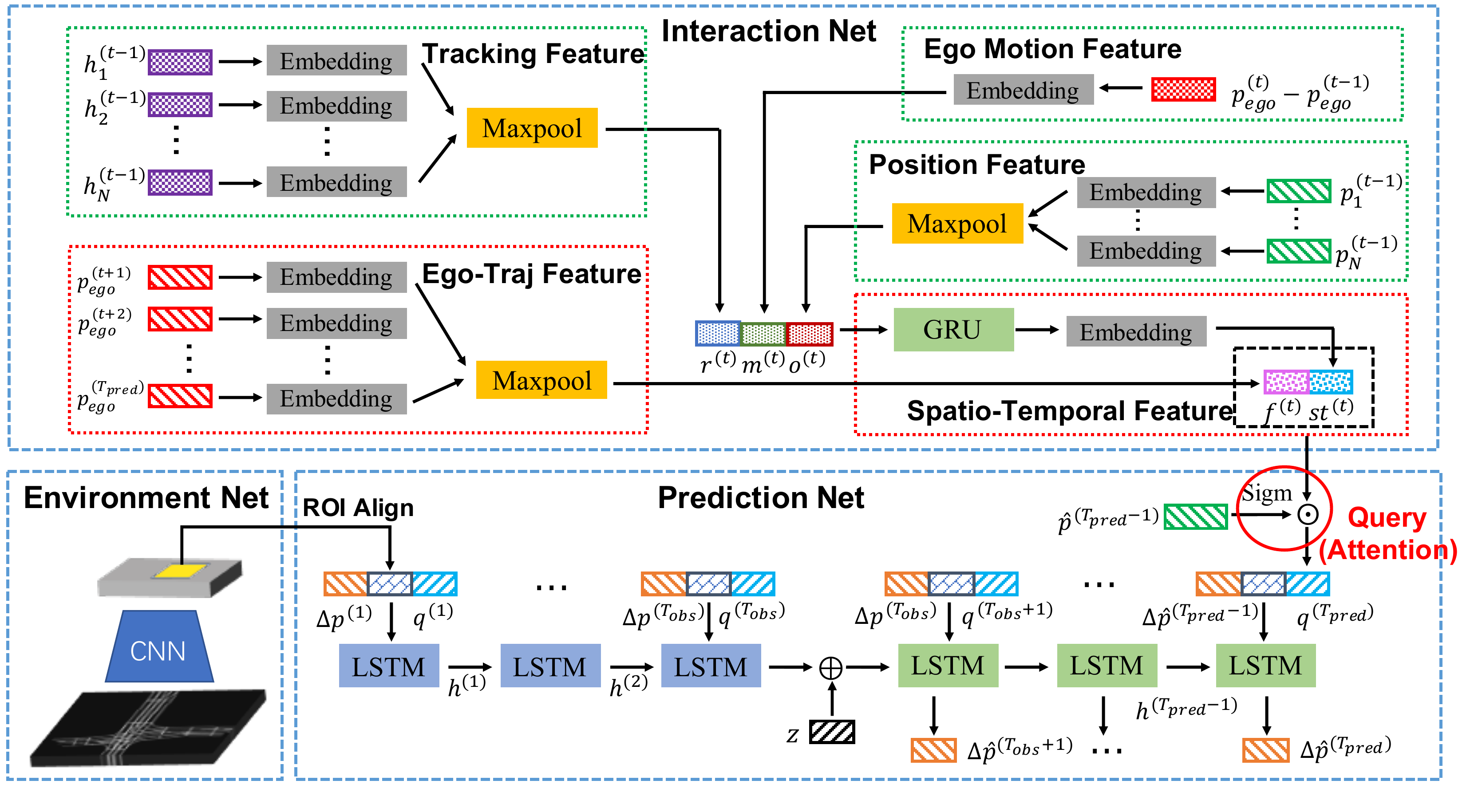}
	\caption{
		Overview of the proposed method.  The proposed method contains three components, a spatio-temporal interaction network, an environment feature extraction network, and a trajectory prediction network.
	}
	\label{fig:2}
\end{figure*}

\section{Related Work}

\subsection{RNN networks and trajectory prediction}

Recurrent neural networks (RNN) and its variants, such as LSTM and GRU, have been shown very successful in many sequence forecasting tasks \cite{LSTMGRU2014}. Therefore, many researches focusing on using RNN and its variants for trajectory prediction. A simple and scalable RNN architecture for human motion prediction is proposed by \cite{simpleRNN2017}. The CIDNN method \cite{CIDNN2018cvpr} uses the inner product of the motion features, which are obtained by LSTMs, to encode the interactions among agents and feeds the interaction features into a multi-layer perceptron for prediction.  By using separate LSTMs for heterogeneous agents on road, the VP-LSTM method \cite{VPLSTM_2019_ICCV} is designed to learn and predict the trajectories of both pedestrians and vehicles simultaneously. In \cite{relation2019ICCV}, a relation gate module is proposed to replace the LSTM unit for capturing a more descriptive spatio-temporal interactions, and both human-human and human-scene interactions from local and global scales are used for future trajectory forecast.  These studies indicate that RNN alone is unable to handle complex scenarios, such as interactions, physical scene and road topology.  Additional structure and operations are always required for accurate, robust and long term prediction.

\subsection{Social behaviors and interactions}

Based on handcraft rules and functions, the social force models \cite{socialForce1995,socialForce2010} use attractive and repulsive forces to describe the interactions of pedestrians in crowd. However, the handcraft rules and functions are unable to generalize for complex interaction scenarios. Instead of handcraft parameters, recent methods use RNN and its variants to learn the parameters directly from data. Social-LSTM \cite{socialLSTM2016} proposes a social pooling layer to model interactions among nearby agents, where the pooling layer uses LSTMs to encode and decode the trajectories.  In \cite{ijcai2017SuForecast},  the method uses LSTMs with social-aware recurrent Gaussian processes to model the complex transitions and uncertainties of agents in crowd.
The SoPhie method \cite{SoPhieCVPR2019} uses the information from both the physical scene context and the social interactions among the agents for prediction.  The TraPHic method \cite{TraPHic2019} proposes to use both the horizon-based and the heterogeneous-based weights to describe interactions among road agents.  A Generative Adversarial Network (GAN) is applied in the social-ways method \cite{socialways2019} to derive plausible future trajectories of agents, where both generator and discriminator networks of the GAN are built by LSTMs.

\subsection{Graph models for trajectory prediction}

Many previous studies formulate interactions of agents as graphs, where nodes refers to the agents, and edges are used to represent the pairwise interactions. Edge weights are used to quantify the importances of the agents to each other. The social-BiGAT method \cite{biGATnips2019} proposes a graph attention network to encode the interactions among humans in a scene and a recurrent encoder-decoder architecture to predict the trajectory.  A dynamic graph-structured model for multimodal trajectories prediction, which is named as the Trajectron, is proposed in \cite{trajectron2019}.  Constructed on a 4-D graph, the TrafficPredict method \cite{TrafficPredict2019} consists of two main layers, an instance layer to learn interactions and a category layer to learn the similarities of instance of the same type. TrafficPredict has shown promising results for trajectory prediction of heterogeneous road agents such as bicycles, vehicles and pedestrians. The STGAT method \cite{STGAT2019cvpr} first uses an LSTM to capture the trajectory information of each agent and applies the graph attention network to model interactions in agents at every time step. Then STGAT adopts another LSTM to learn the temporal correlations for interactions explicitly.

\section{The Proposed Method}

\subsection{Problem Formulation}

In this study, we consider two types of mobile agents: the ego-agent and the other agents. The spatial coordinates of all agents from time step $1$ to $T_{obs}$ are given to predict their future locations from time step $T_{obs+1}$ to $T_{pred}$.
The general formulation of trajectory prediction is represented as,
\ben
\mathbf{Prediction}_{\left\{\theta\right\}}: \left\{\{X_i\}_{i=1}^N, X_{ego}, Y_{ego} \right\}  \longmapsto {\left\{Y_i\right\}}_{i=1}^N ,
\enn
where $X_i$ and  $Y_i$ denote the past and future trajectories of the $i$-th agent, respectively, $X_{ego}$ and $Y_{ego}$ stand for the trajectories of the ego-agent, and $\theta$ denotes the model parameters. Different from previous studies, we consider the prediction problem on the real autonomous driving system where the planned trajectory for the ego-agent $Y_{ego}$ is given for reference. The planned trajectory can improve the prediction accuracy because it brings some prior knowledge on the future. Specifically,  either an observed or a future trajectory can be expressed as a set of temporal coordinates $X_i (\rm{or}\ Y_i)=\left\{\left(x_i^{(1)},y_i^{(1)}\right), \left(x_i^{(2)},y_i^{(2)}\right), \cdots, \left(x_i^{(t)},y_i^{(t)}\right)\right\}$. We use  $p_i^{(t)}=\left(x_i^{(t)},y_i^{(t)}\right)^T$ to denote the location of the $i$-th agent at time step $t$, and set the id of the ego-agent be 0.

As is shown in Fig. \ref{fig:2}, our proposed approach consists of three interrelated components:  an interaction net for spatio-temporal interactive feature extraction, an environment exploration network for decoding dynamic physical scene (i.e., the surrounding road topology), and a trajectory prediction network. Each component and the implementation details of the proposed method is described in details as follows.

\subsection{Interaction Net}

The Agent Interaction Network (AIN) is designed to encode the interaction feature among all the agents in the dynamic scenario. As opposed to the pairwise interactive feature by attention models in previous studies, our method is able to capture the collective influence among the agents. Besides, our method could consider the future movement of the ego-agent for reference. The AIN takes three information sources of all agents as input: the past trajectories, the hidden states of LSTMs and the planned trajectory of the ego-agent. Given these data, AIN computes the global spatio-temporal inter-agents interactions and the future ego-others interaction.

\begin{figure}[htbp]
	\centering
	\includegraphics[width=7cm]{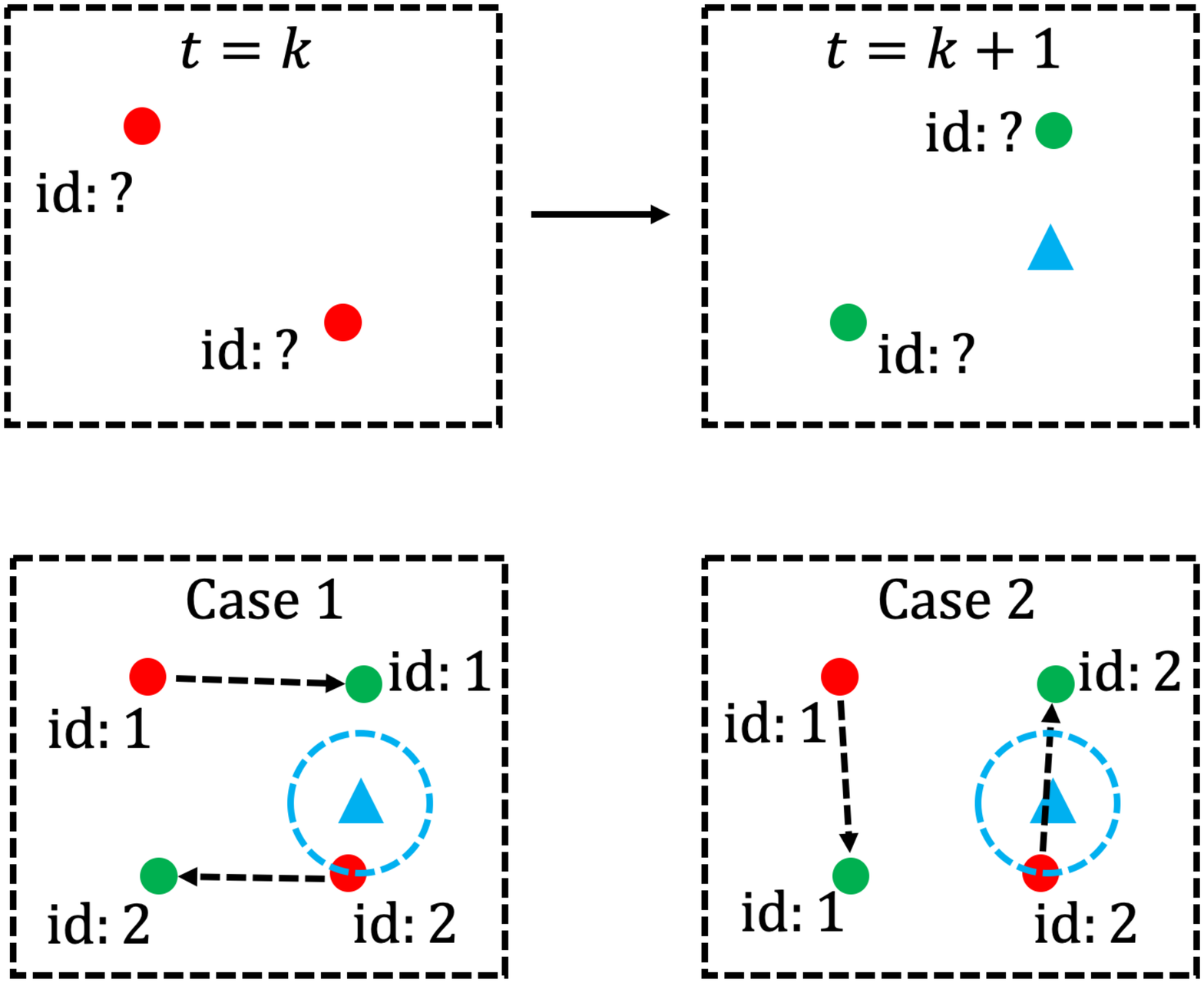}
	\caption{
		An example to show global interaction with or without tracking information. The  top subfigures present the locations of two agents at adjacent time steps without tracking information.  The bottom subfigures show the locations of two agents at adjacent time steps with tracking information.
	}
	\label{fig:1}
\end{figure}

\subsubsection{Global spatio-temporal inter-agents interaction}

The past trajectories of all agents contain the latent patterns of the interactive movement in dynamic scenario. In this module, we intend to learn the latent patterns through a neural network. The learned latent feature represents the global spatio-temporal interaction of all agents on road.

Given the locations of all agents at a time step $t$, we utilize the linear and maxpooling functions to produce the global position feature of size $1 \times d_o$, which is given as below:
\be
e_{o,i}^{(t)} = W_o p_i^{(t)} + b_o, \ \ \forall i \in \left\{0,\dots,N\right\},
\label{(1)}
\\
o^{(t)} = \mathbf{Maxpool}\left(\mathbf{Cat}\left(\left[{e_{o,0}^{(t)}}^T,\cdots,{e_{o,N}^{(t)}}^T\right], 1\right)\right),
\label{(2)}
\en
where $W_o \in \R^{d_o\times 2}$ and $b_o\in \R^{d_o}$ are the weight matrix and bias of the embedding layer. $\mathbf{Cat}\left(\left[\cdot\right],1\right)$ denotes the concatenation function which joints all the inputs along the first dimension. The $\mathbf{Maxpool}\left(\cdot\right)$  function squeezes the spliced data along the same dimension, i.e., the batch dimension.

Moreover, a key problem for the position feature given in Eq. \eqref{(2)} is the temporal issue. Process without temporal information ignores the past interaction and may lead to performance drop. As can be seen in top subfigures of Fig. \ref{fig:1}, the locations of two agents (two circles) at two adjacent time steps are shown.  Without tracking information (the agent id), it is impossible to know which agent and how the agent interacts the blue triangle in the time span. There are two different possible motion behavior of the agents from time step $k$ to $k+1$, as is shown in the bottom subfigures of Fig. \ref{fig:1}.  In case 1, both the two agents could interact with the blue triangle. And in case 2, the blue triangle is more likely to interact with the agent 2. 

To address the temporal issue, we use the hidden states of the LSTMs in the prediction network to track the locations of all agents. The global tracking feature $r^{(t)} \in \mathbb{R}^{1 \times d_r}$ is obtained as follows,
\be
e_{r,i}^{(t)} = W_r h_i^{(t)} + b_r, \ \ \forall i \in \left\{0,\dots,N\right\},
\label{(3)}
\\
r^{(t)} = \mathbf{Maxpool}\left(\mathbf{Cat}\left(\left[{e_{r,0}^{(t)}}^T,\cdots,{e_{r,N}^{(t)}}^T\right], 1\right)\right),
\label{(4)}
\en
where $W_r$ and $b_r$ are the layer parameters, and $h_i^{(t)} = h_{e,i}^{(t)}$ when $t\leq T_{obs}$,  $h_i^{(t)} = h_{h,i}^{(t)}$ when $t\geq T_{obs}+1$.

On real autonomous driving system, one is given the planned trajectory of ego-agent to address the issue of coordinate system transformation (from the world coordinate system to relative coordinate system where the ego-agent centers the origin). From ego-perspective, the global inter-agents interaction module can be mathematically expressed as below:
\be
m^{(t)} = W_m \left(p_0^{(t)} - p_0^{(t-1)}\right)+ b_m,
\label{(5)}
\\
h_{gru}^{(t)} = \mathbf{GRU}\left(\mathbf{Cat}\left(\left[o^{(t)},r^{(t)},m^{(t)}\right], 2\right)\right),
\label{(6)}
\\
st^{(t)} = W_s h_{gru}^{(t)}+ b_s,
\label{(7)}
\en
where the linear layer with parameter $W_m$ and $b_m$ embeds the displacement of the ego-vehicle at two adjacent times into a feature in $\mathbb{R}^{1\times d_m}$. It is worth noting that here we concatenate three kinds of features along the 2nd dimension and produce a comprehensive representation of size  $1 \times d_{st}$. Dimension length $d_{st}$ equals to $\left(d_o+d_r+d_m\right)$. $\mathbf{GRU}$ is similar to LSTM except that it uses less parameters and converge faster with fewer training samples. GRU is used here because the number of ego-agents is much smaller than the number of other mobile agents in our problem.

\subsubsection{Future ego-trajectory interaction}

As the planned trajectory is given, the surrounding agents are inclined to adjust their future motion for collision avoidance. Given ego-trajectory $Y_0$,  we first map it into a high-dimension space using an embedding layer and then pass the obtained embedded feature through a maxpooling function to generate the integrated representation  $f^t \in \mathbb{R}^{1\times d_f}$ of the ego-agent. This representation is what we call a future ego-trajectory feature as it can influence the trajectories of the other on-road agents. The overall process is formulated as follows. 
\ben
e_{f,0}^{(k)} = W_f p_0^{(k)} + b_f, \ \ \forall k, k \in \left[t+1,T_{pred}\right],
\label{(8)}
\\
f^{(t)} = \mathbf{Maxpool}\left(\mathbf{Cat}\left(\left[{e_{f,0}^{(t+1)}}^T,\cdots,{e_{f,0}^{(T_{pred})}}^T\right], 1\right)\right),
\label{(9)}
\enn
Finally, as show in Fig. \ref{fig:2}, the output of the AIN $fst^{(t)} \in \mathbb{R}^{1\times \left(d_{f}+d_{st}\right)}$ is obtained as below:
$$
fst^{(t)} = \mathbf{Cat}\left(\left[f^{(t)}, st^{(t)}\right], 2\right).
\eqno{(10)}
$$

\subsection{Environment Network}

Road topology, such as intersections, turns, and slips, have significant influence on both the speed and directions of the agents. Therefore, it is an important factor in predicting the trajectories of agents. Here we use a network to encode the road topology, where the network is named as the Environment Network (EN). In our method, EN explicitly extracts the drivable area from a High Definition (HD) map. The center lines of the roads are normalized by subtracting the location of the ego-agent for the ego-perspective. And then, we transform the processed lines of roads into a semantic image $\mathbf{I}$ of the map of resolution $H \times W$. That is to say, the ego-agent always locates at the center of the image. Besides, to ensure the consistency of the image and the map, the road areas is trimmed with a fixed size of $h \times w$ meters from the HD map around the ego-agent. Then, the resolution of the semantic image is $\left[h/H, w/W\right]$ meters per pixel. At any time step, EN takes the road image $\mathbf{I}$ as input and encodes the environment through a pre-trained ResNet18 network \cite{resnet18}. The output of the 2nd block of ResNet18 is used as the map feature. Compare to the size of image $\mathbf{I}$, the map feature is downsampled by a factor of 8.

Given the location of an agent, we pool the local road representation at its current location from the computed map feature. The environmental information within $R_s$ meters around the agent is extracted from the obtained map feature. Thus, the corresponding Region Of the Interest (ROI) on the feature maps has a spatial window of $\left[HR_s/4h, WR_s/4w\right]$. We apply ROIAlign on the receptive field to generate a fixed size representation $G_i^{(t)} \in \mathbb{R} ^{C \times K \times K}$, where $C$ is the number of output channels in the last layer, and $K$ is the pooling size. As environment feature $G_i^{(t)}$ is produced,  we feed it to an embedding layer for dimension reduction and feature extraction. The computation of the embedding operation is written as:
\be
g_i^{(t)} = \mathbf{Reshape}\left(G_i^{(t)}, [C\times K \times K, 1]\right),
\label{(11)}
\\
v_i^{(t)} = W_vg_i^{(t)} + b_v,
\label{(12)}
\en
where the function $\mathbf{Reshape}\left(\cdot\right)$ is used to adjust shape size of the target tensor, $W_v$ and $b_v$ are the layer parameters, and the dimension of the $v^{(t)}$ is $d_v$.

\subsection{Trajectory Prediction Network}

\begin{table*}
\centering
\caption{Experimental Results (ADE/FDE) on the ETH and UCY datasets.} \footnotesize
\label{tableETHUCY}
\begin{tabular}{|c||c|c|c|c|c|c|}
\hline
 {\bf Method$\backslash$Dataset}& ETH-univ &  ETH-hotel & UCY-univ & UCY-zara01 &UCY-zara02 & AVG \\ \hline\hline
Linear (Single)    & 1.33/2.94  & {\bf 0.39/0.72}  & 0.82/1.59  & 0.62/1.21     & 0.79/1.59     & 0.79/1.59  \\ \hline
LSTM (Single)    & 1.09/2.41   &0.86/1.91    &   0.61/1.31  & 0.41/0.88    & 0.52/1.11      &0.70/1.52  \\ \hline
Social LSTM      & 1.09/2.35   &0.79/1.76   & 0.67/1.40   &0.47/1.00     & 0.56/1.17      &0.72/1.54  \\ \hline
Social GAN (20V-20) &0.81/1.52 &0.72/1.61 &0.60/1.26   &0.34/0.69    &0.42/0.84      &0.58/1.18 \\ \hline
Social GAN (20VP-20)&0.87/1.62 &0.67/1.37 & 0.76/1.52 &0.35/0.68    &0.42/0.84     &0.61/1.21  \\ \hline
Social Way        & {\bf 0.39/0.64}   & {\bf 0.39/0.66}  &0.55/1.31     &0.44/0.64    &0.51/0.92     &0.45/0.83\\ \hline
Sophie              &0.70/1.43    & 0.76/1.67    & 0.54/1.24   & {\bf 0.30/0.63}   &0.38/0.78    &0.54/1.15\\ \hline
Our method      & {\bf 0.39/0.79}    & 0.51/1.05   & {\bf 0.25/0.56}  & {\bf 0.30/0.61}   & {\bf 0.36/0.73}    & {\bf 0.36/0.75} \\ \hline
\end{tabular}
\end{table*}

The global spaio-temporal interaction and the environment information are encoded by the AIN and the EN, respectively. Besides, given the location of a moving agent, i.e., the $i$-th one, we first compute the local area interaction around the agent through an attention model. This is because an individual always focuses on the surrounding regions as it moves. The attention model is presented below:
\be
e_{c,i}^{(t)} = \sigma \left(W_cp_i^{(t)}+ b_c\right),
\label{(13)}
\\
q_i^{(t)} = \left({fst^{(t)}}\right)^T \odot e_{c,i}^{(t)},
\label{(14)}
\en
where $W_c$ and $b_c$ are the parameters of the embedding layer, $ \sigma \left( \cdot \right)$ is the sigmoid activation function, and $\odot$ is the element-wise vector-vector or matrix-matrix operation.  This layer maps the input into an attention weight $e_{c,i}^{(t)}$ which has the same dimensionality as $fst^{(t)}$.

Following previous works, we utilize an LSTM-based sequence-to-sequence model to solve the prediction problem. For each obstacle, the encoder takes the observed trajectory as input at the first $T_{obs}$ time steps:
$$
e_{p,i}^{(t)} = W_p\left( p_i^{(t)}-p_i^{(t-1)}\right)+ b_p,
\eqno{(15)}
$$
$$
h_{e,i}^{(t)} = \mathbf{LSTM_E}\left(h_{e,i}^{(t-1)}, \left[{e_{p,i}^{(t)}}, {v_i^{(t)}}, {q_i^{(t)}}\right]\right),
\eqno{(16)}
$$
where $W_p$ and $b_p$ are the weights and bias respectively. $h_{e,i}^{(t)}$ is the hidden states of the encoder $\mathbf{LSTM_E}$. Here we set the hidden state to zero vector at the time step 0. The decoding process from time step $T_{obs}+1$ to $T_{pred}$ has a similar flow with the encoding phase, and we use the predicted coordinates as input:
$$
e_{p,i}^{(t)} = W_p\left( \widehat{p}_i^{(t)}-\widehat{p}_i^{(t-1)}\right)+ b_p,
\eqno{(17)}
$$
In practice, we set $\widehat{p}_i^{(T_{obs})}$ to $p_i^{(T_{obs})}$. Our approach forecasts obstacle's position at each prediction moment using the hidden states of the deocder $\mathbf{LSTM_D}$:
$$
h_{d,i}^{(t)} = \mathbf{LSTM_D}\left(h_{d,i}^{(t-1)}, \left[{e_{p,i}^{(t)}}, {v_i^{(t)}}, {q_i^{(t)}}\right]\right),
\eqno{(18)}
$$
$$
\Delta \widehat{p}_i^{{(t+1)}} = W_uh_{d,i}^{(t)} + b_u,
\eqno{(19)}
$$
$$
\widehat{p}_i^{(t+1)} = \widehat{p}_i^{(t)} + \Delta \widehat{p}_i^{t+1}.
\eqno{(20)}
$$

Furthermore, to capture the multi-modal distribution of the movement and increase the robustness of the proposed method, we introduce a gaussian random noise into the decoder to generate multiple plausible trajectories. Specifically, we initialize the hidden state of the $\mathbf{LSTM_D}$ using the last state of the $\mathbf{LSTM_E}$:
$$
e_{h,i}^{(T_{obs})} = \mathbf{Cat}\left(\left[h_{e,i}^{(T_{obs})}, z\right], 1\right),
\eqno{(21)}
$$
$$
h_{d,i}^{(T_{obs})} = W_\phi e_{h,i}^{(T_{obs})} +b_\phi,
\eqno{(22)}
$$
where $z$ is some gaussian random noise. 

\subsection{Implementation Details}

Our network is trained end-to-end by minimizing the mean square error as below:
$$
\mathcal{L} = \frac{1}{NT} \min_{i\in \{1...H\}}  \sum_{j=1}^{N} \sum_{t=T_{obs}+1}^{T_{pred}} \left(\widehat{p}_{j,i}^{(t)}-p_{j}^{(t)} \right)^{2}, \eqno{(23)}
$$
where $T$ is the prediction time steps which equals $T_{pred}-T_{obs}$. $H$ is the number of modalities (predicted trajectories). We only back propagate the gradient to the modality with the minimum error.

We set the output dimensions of all the embedding layers (exclude attention and noise embedding layer in Trajectory Prediction Network) to be 64. The GRU in the AIN has 128 cells, while the LSTMs in the Trajectory Prediction Network have 64 cells. In the EN, the local area size $R_s$ and the pool size $K$ are set as 20 and 3 respectively. Meanwhile, both the height $H$ and width $W$ of the road semantic image are set to 224. The road area size $h$ and $w$ are set to be 100 meters. Our network is trained with a batch size of 8 for 20000 steps using Adam optimizer with an initial learning rate of 0.0005. The entire training process is finished in the platform with an NVIDIA GeForce RTX2080 GPU.

\begin{table*}
\centering
\caption{Experimental Results (ADE/FDE) on the  ApolloScape Dataset.} \footnotesize
\label{tableApollo}
\begin{tabular}{|c||c|c|c|c|c|c|c|}
\hline
 {\bf Dataset$\backslash$Method}&  Linear &  KF & LSTM & Noise-LSTM & Social LSTM &  Social GAN & Our method \\ \hline\hline
ApolloScape   & 1.95/3.39 &2.48/4.33 & 1.63/2.85 & 1.49/2.58 & 1.23/2.11 & 1.21/2.14 &  {\bf 1.11/1.91 }  \\ \hline
\end{tabular}
\end{table*}

\begin{table*}
\centering
\caption{Experimental Results (ADE/FDE) on the  Argoverse Dataset.} \footnotesize
\label{tableArgo}
\begin{tabular}{|c||c|c|c|c|c|c|c|}
\hline
 {\bf Dataset$\backslash$Method}&  Linear &  KF & LSTM & Noise-LSTM & Social LSTM &  Social GAN & Our method \\ \hline\hline
Relative coordinate   & 1.84/3.91. & 2.53/5.84 & 1.47/3.18 & 1.48/3.17 & 1.23/2.49 & 1.31/2.63 & {\bf 1.18/2.45}  \\ \hline
World coordinate      &  1.57/3.31 & 2.56/5.96 & 1.24/2.63 & 1.25/2.61 & 1.22/2.45 & 1.32/2.67 & {\bf 1.15/2.29} \\ \hline
\end{tabular}
\end{table*}

\section{Experiments}

In this section, we evaluate the proposed method on four benchmark datasets for future trajectory prediction and demonstrate our method performs favorably against state-of-the-art prediction methods. The codes and pre-trained models of our method will be released to the public.

\subsection{Datasets Description}

ETH \cite{ETH2009} and UCY \cite{UCY2007} are two common benchmarks for pedestrian trajectory prediction. These two datasets consists of 5 scenes, including ETH-univ, ETH-hotel, UCY-zara01, UCY-zara02 and UCY-univ. There are totally 1536 pedestrians in total with thousands of nonlinear trajectories. The same leave-one-set-out strategy as in previous study \cite{socialLSTM2016} is used to evaluate the compared methods.  

Besides pedestrian walking datasets, the ApolloScape \cite{TrafficPredict2019} and the Argoverse \cite{Argoverse2019} datasets are utilized to demonstrate the performance of the compared methods.  ApolloScape dataset is comprised of different kinds of traffic agents which include cars, buses, pedestrians and bicycles.  This dataset is very challenging because it is a heterogeneous multi-agent system. On the other hand, Argoverse dataset contains 327790 sequences of different scenarios. Each sequence follows the trajectory of ego-agent for 5 seconds while keeping track of all other agents (cars, pedestrians etc.). The dataset is split into a training data with 208272 sequences and a validation data with 79391 sequences. For ApolloScape, the trajectories of  3 seconds (6 time steps) are observed and the prediction methods are required to predict the trajectory in the following 3 seconds (6 time steps). And  for Argoverse, 2 seconds with 20 time steps are observed and the methods are required to predict the trajectories in the following 3 seconds with 30 time steps.

\begin{table}
\caption{Ablation Study on the Argoverse Dataset.}
\label{table:Ablation} \footnotesize
\centering
\begin{tabular}{|c|c|c|c|c|c|c|c|} \hline
\multirow{2}{*}{Method} & \multicolumn{5}{c|}{ Component}  & \multicolumn{2}{c|}{ Metric }  \\
\cline{2-8} & PF               & TF                   & EMF           & ETF         & EF               &ADE & FDE    \\ \hline
 Baseline &         -           & -                     & -             & -            & -              & 1.48 & 3.17 \\
 Our-v1   & $\checkmark$ & -                     & -             & -            &  -             & 1.32 & 2.70\\
 Our-v2   & $\checkmark$ & $\checkmark$  & -             &  -           &   -             & 1.24 & 2.56 \\
 Our-v3   & $\checkmark$ & $\checkmark$  & $\checkmark$  & -    &  -             &1.22  & 2.51\\
 Our-v4   & $\checkmark$ & $\checkmark$  & $\checkmark$  & $\checkmark$ &   -            & 1.19 & 2.47 \\
 Our-full & $\checkmark$ & $\checkmark$   & $\checkmark$  & $\checkmark$ & $\checkmark$   & {\bf 1.18}& \bf{2.45} \\ \hline
\end{tabular}
\end{table}

\subsection{Experimental setup}

The experimental results are reported in terms of two evaluation metrics, the Average Displacement Error (ADE) and Final Displacement Error (FDE). ADE is defined as the mean square error over all prediction points of a trajectory and the ground truth points of the trajectory, whereas FDE is the distance between the predicted final location and the ground truth final location at the end of the prediction time period. 

We use the the linear regressor,  the extended Kalman Filter (KF), and the vanila-LSTM as the baselines. Moreover, many state-of-the-art trajectory prediction methods are compared. Social-LSTM \cite{socialLSTM2016} is a prediction method that combines LSTMs with a social pooling layer.  Social-GAN \cite{SocialGAN2018} applies a GAN model to social LSTMs for prediction. Social-Way \cite{socialways2019} utilizes a GAN model to propose plausible future trajectories and train the predictor. Sophie \cite{SoPhieCVPR2019} introduces a social and physical attention mechanism to a GAN predictor.

Because there is no HD map information and the planned trajectory for the ego-agent, there is no ego-trajectory feature, ego-motion feature and environment feature in the proposed method for the ETH $\&$ UCY, and the ApolloScape datasets.  For Argoverse dataset, the proposed method with all the features and components is implemented for comparisons.

\subsection{Performance Evaluation}

{\bf ETH $\&$ UCY } The experimental results on the ETH and UCY datasets are presented in Table \ref{tableETHUCY}. As expected, the baselines, the Linear and LSTM, are incapable in capturing the complexity patterns in the trajectories of pedestrians.  Our method outperforms the other methods on the UCY-univ and UCY-zara02 subsets and shows competitive results on the ETH-univ and UCY-zara01 subsets.  On the ETH-hotel, both linear and social way methods show lower prediction errors than other methods.  This indicate that the trajectories in ETH-hotel are linearly distributed and thus are simpler than the other 4 subsets. As the methods are all trained on the other 4 subsets, these nonlinear predictors, such as Social LSTM, Social-GAN, Sophie, show poor generalization ability on ETH-hotel.  On the other hand, our method still outperforms Social LSTM, Social-GAN, and Sophie on the ETH-hotel subset.

{\bf AplloScape }  The performance of the compared methods on the ApolloScape is shown in Table \ref{tableApollo}.  As can be seen, the proposed method outperforms the runner-up in term of ADE/FDE with about 10\% improvement in accuracy. It means that our interaction net has faithfully learn the intrinsic interactive patterns and the attention module could extract the specialized feature for each category of the heterogeneous traffic-agents 

{\bf Argoverse } Argoverse provides the HD road map and the planned path for ego-car. The proposed method with all the components is implemented. The experimental results are shown in Table \ref{tableArgo}. As can be seen, the prediction error of the proposed method is significantly lower than the other methods. We observe 11\% and 4\% improvement in ADE comparing the Social GAN and Social LSTM methods when the relative coordinate system (ego-perspective) is used. The improvement in ADE is 14\% and 6\% comparing the Social GAN and the social LSTM methods when the world coordinate system is used.  

{\bf Ablation Study} The proposed method is comprised of multiple separate components, each with different functions. To show the effectiveness of the components, we perform ablation study of the components of the proposed method and the results are presented in Table \ref{table:Ablation}.  We use PF, TF, EMF, ETF, EF to represent the position feature, tracking feature, ego motion feature, ego trajectory feature, and environment feature of the proposed method, respectively. The results indicate that PF and TF are contributive to the significant improvement, and the values of the reduction in prediction error  are 0.47 and 0.14, respectively.  And EMF, ETF, EF all show some level of contributions such that the values of the reduction in prediction error are 0.5, 0.4 and 0.2, respectively.  The value of the reduction in error is dropping because it is harder to reduce the prediction error when the error is lower. 

\section{Conclusion}

We propose a new method for trajectory forecasting of multiple agents in dynamic scenes. The method is able to extract the global spatio-temporal interaction feature from the past trajectories, and consider the temporal interactions among agents by soft tracking. An environment net is introduced in our method to encode the road topology for accurate prediction. And the prediction net combines the features of spatio-temporal interactions and environment to prediction the future trajectories of agents. Experiments on four benchmark datasets are presented and the ablation study is implemented to show the effectiveness of each component of the method. 

\bibliographystyle{named}
{ \small
\bibliography{ijcai20}

\begin{thebibliography}{}

\bibitem[\protect\citeauthoryear{Alahi \bgroup \em et al.\egroup
  }{2016}]{socialLSTM2016}
A.~Alahi, K.~Goel, V.~Ramanathan, A.~Robicquet, F.~Li, and S.~Savarese.
\newblock Social lstm: Human trajectory prediction in crowded spaces.
\newblock In {\em The IEEE Conference on Computer Vision and Pattern
  Recognition (CVPR)}, June 2016.

\bibitem[\protect\citeauthoryear{Amirian \bgroup \em et al.\egroup
  }{2019}]{socialways2019}
J.~Amirian, J.~B. Hayet, and J.~Pettre.
\newblock Social ways: Learning multi-modal distributions of pedestrian
  trajectories with gans.
\newblock In {\em The IEEE Conference on Computer Vision and Pattern
  Recognition (CVPR) Workshops}, June 2019.

\bibitem[\protect\citeauthoryear{Bi \bgroup \em et al.\egroup
  }{2019}]{VPLSTM_2019_ICCV}
Huikun Bi, Zhong Fang, Tianlu Mao, Zhaoqi Wang, and Zhigang Deng.
\newblock Joint prediction for kinematic trajectories in
  vehicle-pedestrian-mixed scenes.
\newblock In {\em The IEEE International Conference on Computer Vision (ICCV)},
  October 2019.

\bibitem[\protect\citeauthoryear{Chandra \bgroup \em et al.\egroup
  }{2019}]{TraPHic2019}
R.~Chandra, A.~Bhattacharya, U.and~Bera, and D.~Manocha.
\newblock Traphic: Trajectory prediction in dense and heterogeneous traffic
  using weighted interactions.
\newblock In {\em The IEEE Conference on Computer Vision and Pattern
  Recognition (CVPR)}, June 2019.

\bibitem[\protect\citeauthoryear{Chang \bgroup \em et al.\egroup
  }{2019}]{Argoverse2019}
M.~Chang, J.~W. Lambert, P.~Sangkloy, J.~Singh, S.~Bak, A.~Hartnett, D.~Wang,
  P.~Carr, S.~Lucey, D.~Ramanan, and J.~Hays.
\newblock Argoverse: 3d tracking and forecasting with rich maps.
\newblock In {\em Conference on Computer Vision and Pattern Recognition
  (CVPR)}, 2019.

\bibitem[\protect\citeauthoryear{Choi and Dariush}{2019}]{relation2019ICCV}
C.~Choi and B.~Dariush.
\newblock Looking to relations for future trajectory forecast.
\newblock In {\em The IEEE International Conference on Computer Vision (ICCV)},
  October 2019.

\bibitem[\protect\citeauthoryear{Chung \bgroup \em et al.\egroup
  }{2014}]{LSTMGRU2014}
J.~Chung, C.~Gulcehre, K.H. Cho, and Y.~Bengio.
\newblock Empirical evaluation of gated recurrent neural networks on sequence
  modeling, 2014.

\bibitem[\protect\citeauthoryear{Gupta \bgroup \em et al.\egroup
  }{2018}]{SocialGAN2018}
Agrim Gupta, Justin Johnson, Li~Fei-Fei, Silvio Savarese, and Alexandre Alahi.
\newblock Social gan: Socially acceptable trajectories with generative
  adversarial networks.
\newblock In {\em The IEEE Conference on Computer Vision and Pattern
  Recognition (CVPR)}, June 2018.

\bibitem[\protect\citeauthoryear{{He} \bgroup \em et al.\egroup
  }{2016}]{resnet18}
K.~{He}, X.~{Zhang}, S.~{Ren}, and J.~{Sun}.
\newblock Deep residual learning for image recognition.
\newblock In {\em 2016 IEEE Conference on Computer Vision and Pattern
  Recognition (CVPR)}, pages 770--778, June 2016.

\bibitem[\protect\citeauthoryear{Helbing and Moln\'ar}{1995}]{socialForce1995}
D.~Helbing and P.~Moln\'ar.
\newblock Social force model for pedestrian dynamics.
\newblock {\em Phys. Rev. E}, 51:4282--4286, May 1995.

\bibitem[\protect\citeauthoryear{Huang \bgroup \em et al.\egroup
  }{2019}]{STGAT2019cvpr}
Y.~Huang, H.~Bi, Z.~Li, T.~Mao, and Z.~Wang.
\newblock Stgat: Modeling spatial-temporal interactions for human trajectory
  prediction.
\newblock In {\em The IEEE International Conference on Computer Vision (ICCV)},
  October 2019.

\bibitem[\protect\citeauthoryear{Ivanovic and Pavone}{2019}]{trajectron2019}
Boris Ivanovic and Marco Pavone.
\newblock The trajectron: Probabilistic multi-agent trajectory modeling with
  dynamic spatiotemporal graphs.
\newblock In {\em The IEEE International Conference on Computer Vision (ICCV)},
  October 2019.

\bibitem[\protect\citeauthoryear{Kosaraju \bgroup \em et al.\egroup
  }{2019}]{biGATnips2019}
Vineet Kosaraju, Amir Sadeghian, Roberto Mart\'{\i}n-Mart\'{\i}n, Ian Reid,
  Hamid Rezatofighi, and Silvio Savarese.
\newblock Social-bigat: Multimodal trajectory forecasting using bicycle-gan and
  graph attention networks.
\newblock In {\em Advances in Neural Information Processing Systems 32}, pages
  137--146. 2019.

\bibitem[\protect\citeauthoryear{{Lefèvre} \bgroup \em et al.\egroup
  }{2011}]{bayesian2011}
S.~{Lefèvre}, C.~{Laugier}, and J.~{Ibañez-Guzmán}.
\newblock Exploiting map information for driver intention estimation at road
  intersections.
\newblock In {\em 2011 IEEE Intelligent Vehicles Symposium (IV)}, pages
  583--588, June 2011.

\bibitem[\protect\citeauthoryear{Lerner \bgroup \em et al.\egroup
  }{2007}]{UCY2007}
A.~Lerner, Y.~Chrysanthou, and D.~Lischinski.
\newblock Crowds by example.
\newblock {\em Computer Graphics Forum}, 26(3):655--664, 2007.

\bibitem[\protect\citeauthoryear{Liang \bgroup \em et al.\egroup
  }{2019}]{peekingfuture2019}
Junwei Liang, Lu~Jiang, Juan~Carlos Niebles, Alexander~G. Hauptmann, and
  Li~Fei-Fei.
\newblock Peeking into the future: Predicting future person activities and
  locations in videos.
\newblock In {\em The IEEE Conference on Computer Vision and Pattern
  Recognition (CVPR)}, June 2019.

\bibitem[\protect\citeauthoryear{Ma \bgroup \em et al.\egroup
  }{2019a}]{TrafficPredict2019}
Y.~Ma, X.~Zhu, S.~Zhang, R.~Yang, W.~Wang, and D.~Manocha.
\newblock Trafficpredict: Trajectory prediction for heterogeneous
  traffic-agents.
\newblock {\em Proceedings of the AAAI Conference on Artificial Intelligence},
  33:6120--6127, 07 2019.

\bibitem[\protect\citeauthoryear{Ma \bgroup \em et al.\egroup
  }{2019b}]{baiduAAAI2019}
Yuexin Ma, Xinge Zhu, Sibo Zhang, Ruigang Yang, Wenping Wang, and Dinesh
  Manocha.
\newblock Trafficpredict: Trajectory prediction for heterogeneous
  traffic-agents.
\newblock In {\em The AAAI Conference on Artificial Intelligence (AAAI)}, pages
  6120--6127, February 2019.

\bibitem[\protect\citeauthoryear{{Martinez} \bgroup \em et al.\egroup
  }{2017}]{simpleRNN2017}
J.~{Martinez}, M.~J. {Black}, and J.~{Romero}.
\newblock On human motion prediction using recurrent neural networks.
\newblock In {\em 2017 IEEE Conference on Computer Vision and Pattern
  Recognition (CVPR)}, pages 4674--4683, July 2017.

\bibitem[\protect\citeauthoryear{{Pellegrini} \bgroup \em et al.\egroup
  }{2009}]{ETH2009}
S.~{Pellegrini}, A.~{Ess}, K.~{Schindler}, and L.~{van Gool}.
\newblock You'll never walk alone: Modeling social behavior for multi-target
  tracking.
\newblock In {\em 2009 IEEE 12th International Conference on Computer Vision},
  pages 261--268, Sep. 2009.

\bibitem[\protect\citeauthoryear{Pellegrini \bgroup \em et al.\egroup
  }{2010}]{socialForce2010}
S.~Pellegrini, A.~Ess, and L.~Van~Gool.
\newblock Improving data association by joint modeling of pedestrian
  trajectories and groupings.
\newblock In Kostas Daniilidis, Petros Maragos, and Nikos Paragios, editors,
  {\em European Conference on Computer Vision (ECCV)}, pages 452--465, Berlin,
  Heidelberg, 2010. Springer Berlin Heidelberg.

\bibitem[\protect\citeauthoryear{Rasmussen and Williams}{2005}]{GP2005}
C.~E. Rasmussen and C.~K.~I. Williams.
\newblock {\em Gaussian Processes for Machine Learning (Adaptive Computation
  and Machine Learning)}.
\newblock The MIT Press, Cambridge, MA, USA, 2005.

\bibitem[\protect\citeauthoryear{Sadeghian \bgroup \em et al.\egroup
  }{2019}]{SoPhieCVPR2019}
Amir Sadeghian, Vineet Kosaraju, Ali Sadeghian, Noriaki Hirose, Hamid
  Rezatofighi, and Silvio Savarese.
\newblock Sophie: An attentive gan for predicting paths compliant to social and
  physical constraints.
\newblock In {\em The IEEE Conference on Computer Vision and Pattern
  Recognition (CVPR)}, June 2019.

\bibitem[\protect\citeauthoryear{Su \bgroup \em et al.\egroup
  }{2017}]{ijcai2017SuForecast}
H.~Su, J.~Zhu, Y.~Dong, and B.~Zhang.
\newblock Forecast the plausible paths in crowd scenes.
\newblock In {\em Proceedings of the Twenty-Sixth International Joint
  Conference on Artificial Intelligence, {IJCAI-17}}, pages 2772--2778, 2017.

\bibitem[\protect\citeauthoryear{{Toledo-Moreo} and
  {Zamora-Izquierdo}}{2009}]{IMM2009}
R.~{Toledo-Moreo} and M.~A. {Zamora-Izquierdo}.
\newblock Imm-based lane-change prediction in highways with low-cost gps/ins.
\newblock {\em IEEE Transactions on Intelligent Transportation Systems},
  10(1):180--185, March 2009.

\bibitem[\protect\citeauthoryear{Van~den Berg \bgroup \em et al.\egroup
  }{2011}]{nbody2011}
Jur Van~den Berg, Stephen~J. Guy, Ming Lin, and Dinesh Manocha.
\newblock Reciprocal n-body collision avoidance.
\newblock In C{\'e}dric Pradalier, Roland Siegwart, and Gerhard Hirzinger,
  editors, {\em Robotics Research}, pages 3--19, Berlin, Heidelberg, 2011.
  Springer Berlin Heidelberg.

\bibitem[\protect\citeauthoryear{Xu \bgroup \em et al.\egroup
  }{2018}]{CIDNN2018cvpr}
Y.~Xu, Z.~Piao, and S.~Gao.
\newblock Encoding crowd interaction with deep neural network for pedestrian
  trajectory prediction.
\newblock In {\em The IEEE Conference on Computer Vision and Pattern
  Recognition (CVPR)}, June 2018.

\end{thebibliography}
}
\end{document}